\documentclass[letterpaper, 10 pt, conference]{ieeeconf}
\IEEEoverridecommandlockouts
\usepackage[utf8]{inputenc} 
\usepackage[T1]{fontenc}    
\usepackage{hyperref}       
\usepackage{url}            
\usepackage{booktabs}       
\usepackage{amsfonts}       
\usepackage{nicefrac}       
\usepackage{microtype}      
\usepackage{xcolor}         

\title{ \LARGE \bf
Humanoid Goalkeeper: Learning from Position Conditioned  \\ Task-Motion Constraints
\vspace{-0.5cm}
}
\author{
\authorblockN{
Junli Ren $^{*,1,2}$ \quad
Junfeng Long$^{*,2}$ \quad
Tao Huang$^{2}$ \quad
Huayi Wang$^{2}$ \quad
Zirui Wang$^{2}$ \quad
Feiyu Jia$^{2}$ \quad \\
Wentao Zhang$^{2}$ \quad 
Jingbo Wang$^{2,\dag}$ \quad
Ping Luo$^{1,2,\dag}$ \quad
Jiangmiao Pang$^{2,\dag}$
}
\authorblockA{
\textsuperscript{1}The University of Hong Kong \quad
\textsuperscript{2}Shanghai AI Laboratory \quad 
\textsuperscript{$*$}Equal Contribution \quad
\textsuperscript{$\dagger$}Equal Advising\\
}
}

\usepackage{amsmath}       
\usepackage{amssymb}       
\usepackage{amsfonts}      
\usepackage{mathtools}     
\usepackage{bm}            
\usepackage{bbm}           
\usepackage{xfrac}
\usepackage{times}         
\usepackage{dsfont}        
\usepackage{xspace}        
\usepackage{soul}
\setul{1.5pt}{1pt}

\usepackage{graphicx}      
\usepackage{wrapfig}       
\usepackage{float}         
\usepackage{stfloats}      
\usepackage{capt-of}       
\usepackage[font=footnotesize,labelfont=bf]{caption}      
\usepackage{subfigure}

\usepackage{booktabs}      
\usepackage{multirow}      
\usepackage[flushleft]{threeparttable} 
\usepackage{arydshln}      

\usepackage[dvipsnames,table,xcdraw]{xcolor} 

\definecolor{mydarkblue}{rgb}{0,0.08,0.45}
\definecolor{mydarkgreen}{RGB}{0, 139, 69}
\definecolor{mygreen2}{RGB}{0, 205, 0}
\definecolor{mybrown}{RGB}{139, 69, 19}
\definecolor{boxblue}{RGB}{79,173,234}
\definecolor{tablepeach}{RGB}{255, 240, 235}
\definecolor{tablepurple}{RGB}{248,235,252}
\definecolor{tableblue}{RGB}{235,241,255}

\usepackage{hyperref}                   
\usepackage[capitalise, nameinlink]{cleveref} 
\usepackage{xurl}                       
\definecolor{citecolor}{HTML}{c03d3e}

\hypersetup{
    colorlinks=true,
    linkcolor=mydarkgreen,
    urlcolor=mydarkgreen,
    citecolor=mydarkgreen,
}

\usepackage[linesnumbered,ruled,vlined]{algorithm2e} 

\usepackage{pifont}  
\usepackage{dsfont}  
\usepackage{lipsum}  
\usepackage{siunitx} 
\usepackage{multicol} 
\let\labelindent\relax
\usepackage{enumitem}

\usepackage[
    style=ieee,
    natbib=true,
    citestyle=numeric-comp,
    doi=false,
    isbn=false,
    url=false
]{biblatex} 
\usepackage{etoolbox} 
\AtBeginBibliography{\small} 
\begin{document}
\twocolumn[{\renewcommand\twocolumn[1][]{#1}
\maketitle
\vspace{-0.5cm}
\begin{center}
    \centering
    \captionsetup{type=figure}
     \includegraphics[width=1.0\textwidth]{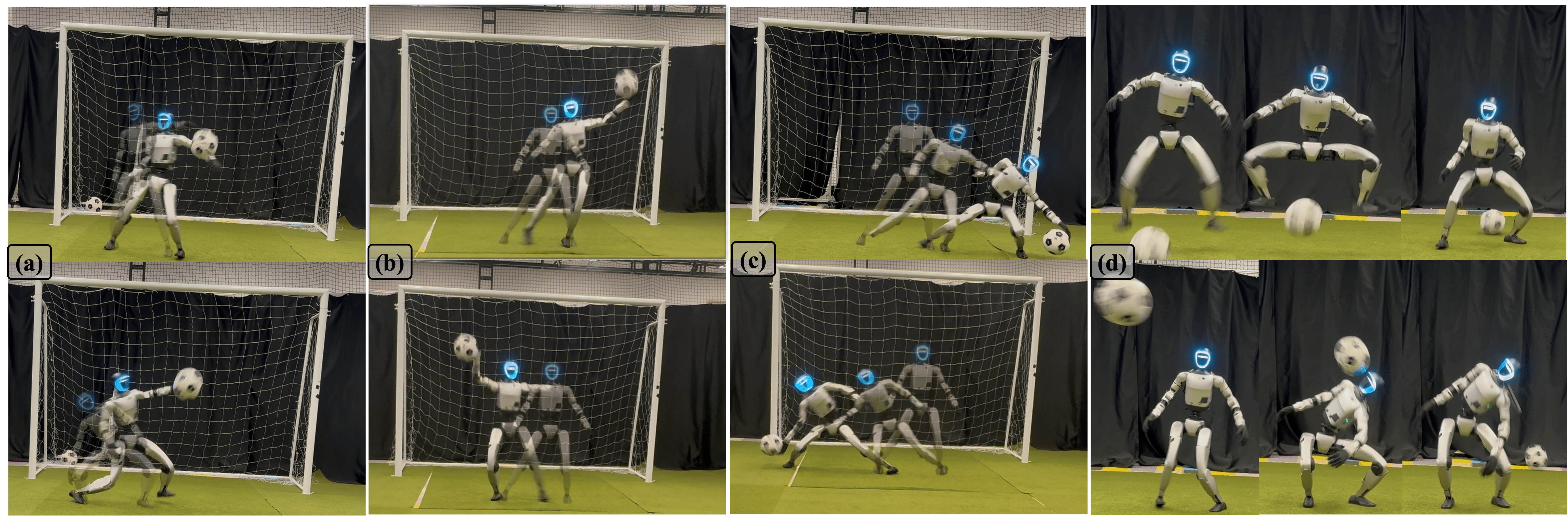}
     \vspace{-0.17in}
    \caption{ We present \textbf{Humanoid Goalkeeper}, capable of performing goalkeeping tasks across various regions with a wide operational range \textcolor{mydarkgreen}{(b, c)}, and initiating from arbitrary postures \textcolor{mydarkgreen}{(a)}. Our policy enables the robot to escape an incoming ball with jump and squat motions \textcolor{mydarkgreen}{(d)}.}
    \label{fig:teaser}
\end{center}
\vspace{-0.0in}
}]

\begin{abstract}
We present a reinforcement learning framework for autonomous goalkeeping with humanoid robots in real-world scenarios. While prior work has demonstrated similar capabilities on quadrupedal platforms, humanoid goalkeeping introduces two critical challenges: (1) generating natural, human-like whole-body motions, and (2) covering a wider guarding range with an equivalent response time. Unlike existing approaches that rely on separate teleoperation or fixed motion tracking for whole-body control, our method learns a single end-to-end RL policy, enabling fully autonomous, highly dynamic, and human-like robot-object interactions. To achieve this, we integrate multiple human motion priors conditioned on perceptual inputs into the RL training via an adversarial scheme. We demonstrate the effectiveness of our method through real-world experiments, where the humanoid robot successfully performs agile, autonomous, and naturalistic interceptions of fast-moving balls. In addition to goalkeeping, we demonstrate the generalization of our approach through tasks such as ball escaping and grabbing. Our work presents a practical and scalable solution for enabling highly dynamic interactions between robots and moving objects, advancing the field toward more adaptive and lifelike robotic behaviors. More details, including videos and implementation, are available on our \href{https://humanoid-goalkeeper.github.io/Goalkeeper/}{\textbf{project website}} 
and \href{https://github.com/InternRobotics/Humanoid-Goalkeeper}{\textbf{GitHub repository}}.
\end{abstract}      
\section{Introduction}

Developing robotic soccer capabilities inherently requires tight integration of perception, decision-making, and agile motor control, making it an appealing benchmark \cite{Haarnoja2023LearningAS, Beukman2024RobocupGymAC, Tirumala2024LearningRS, su2025toward, fernandez2025clapclusteringlocalizen} for evaluating motion intelligence. Within this domain, goalkeeping stands out as a distinct subtask with many underexplored challenges. Unlike dribbling \cite{Ji2023DribbleBotDL, wang2025dribble} or shooting \cite{Ji2022HierarchicalRL, Marew2024ABA}, which primarily involve the lower limbs and interact with slow or static objects, goalkeeping requires rapid, full-body responses to highly dynamic stimuli. Advancing robotic goalkeeping is therefore a crucial step toward fully autonomous, real-time physical intelligence.

Despite the growing interest in robotic goalkeeping, including recent advances in dynamic skills on quadrupeds, existing systems remain limited in several critical aspects: being confined to kid-sized platforms \cite{booster2025t1}, relying on fixed motion primitives \cite{Ze2025TWISTTW} and operating within narrow interception ranges. Meanwhile, emerging developments in reinforcement learning (RL) \cite{schulman2017proximal,rudin2022learning}, imitation-driven motion priors \cite{peng2021amp, Tang2023HumanMimicLN}, and human motion recovery pipelines \cite{shen2024world, Zhang2025HumanMMGH, Shen2024WorldGroundedHM} open new avenues for addressing these limitations. Coupled with the advent of high-performance humanoid hardware and scalable learning frameworks, it is now feasible to explore goalkeeper behaviors that: (1) exhibit human-like, whole-body interactive motions; (2) respond effectively under extreme time constraints; and (3) perform highly dynamic actions to cover broad and diverse interception regions.

In this paper, we present the Humanoid Goalkeeper, a reinforcement learning–based framework for agile and autonomous goalkeeping on humanoid robots. Our method learns a single end-to-end control policy that integrates task-specific rewards with motion priors through an adversarial training scheme, enabling policies that optimize both task performance and motion realism. To address the challenge of wide operational coverage, we condition motion learning on ball landing positions, effectively dividing the constraint space. To ensure real-world feasibility, we incorporate perception noise, trajectory estimation, and multi-modal sensing into the training loop to close the sim-to-real gap. Through extensive evaluations in both simulation and hardware, we demonstrate that our approach enables fast, reactive, and naturalistic interception of high-speed flying balls. Furthermore, we show that the learned policy generalizes to related dynamic interaction tasks such as escaping and grabbing, highlighting the framework’s potential for broader lifelike and adaptive humanoid behaviors.

\section{related work}

\textbf{Dynamic Object Interactions with floating base.} Recent advances have yielded impressive results in interactions between dynamic object and mobile robots, with notable applications in tasks such as badminton \cite{ma2025learning, Wang2025IntegratingLM}, table tennis \cite{su2025hitterhumanoidtabletennis, DAmbrosio2024AchievingHL, Lee2024LearningDR, hu2025towards, nguyen2025whole}, mobile-manipulator-based catching \cite{Zhang2024CatchIL, Abeyruwan2023AgileCW}, throwing \cite{Ma2025LearningAW}, and soccer \cite{huang2023creating, su2025toward, hou2025localization, Ji2023DribbleBotDL, wang2025dribble, Haarnoja2023LearningAS}. These studies have successfully demonstrated millisecond-level interactions between robots and dynamic objects, with a strong integration of perception, planning, and control systems. These methods excel in specific domains such as performing consecutive actions \cite{su2025hitterhumanoidtabletennis}, multi-agent coordination \cite{Xu2025VolleyBotsAT, su2025toward} or professional level skill \cite{DAmbrosio2024AchievingHL}, but they are constrained by narrow movement ranges or lack the dynamic whole-body agility required for humanoid systems. In this work, we address these gaps by developing a humanoid goalkeeper capable of covering wide movement ranges, while executing agile and human-like whole-body motions.

\textbf{Whole body control with motion priors} 
Integrating imitation learning with the RL framework has become a popular approach for transferring human motion skills to humanoid robots. A majority of these methods strictly mimic reference trajectories following the fixed time phase \cite{Chen2025GMTGM, He2025ASAPAS, Liao2025BeyondMimicFM, Allshire2025VisualIE,  He2024OmniH2OUA}. These methods are typically limited to non-interactive tasks, such as dancing or locomotion \cite{Xie2025KungfuBotPH, Ji2024ExBody2AE}, or are restricted to predefined interaction moments within the reference \cite{su2025hitterhumanoidtabletennis, Liu2024MimickingBenchAB, Xue2025LeVERBHW, yang2025omniretarget, yin2025visualmimic, weng2025hdmilearninginteractivehumanoid}. In contrast, adversarial motion priors (AMP) \cite{peng2021amp} offer an alternative approach by integrating expert data without enforcing strict temporal alignment. However, this potential has yet to be fully explored in real-world humanoid tasks, with current frameworks limited to simpler locomotion tasks \cite{Lin2025HWCLocoAH, Ma2025StyleLocoGA}. Meanwhile, multi-motion prior synthesis has been explored through task-condition tokens \cite{Pan2025TokenHSIUS} or sub-task policy selection \cite{Wang2024SkillMimicLB} in the animation domain. Similarly, previous quadruped goalkeepers have learned to select from a set of pre-learned low-level skills \cite{huang2023creating} for different ball target regions. Building upon these experiences, our approach trains the robot with task rewards and motion discriminators specific to different ball target regions, activating only the corresponding constraints during each trial. The proposed pipeline results in a one-stage, end-to-end, hardware-deployable policy that achieves dynamic interactions while maintaining reference motion resemblance.

\section{Method}
\begin{figure}[htbp]
  \centering
  \includegraphics[width=0.48 \textwidth]{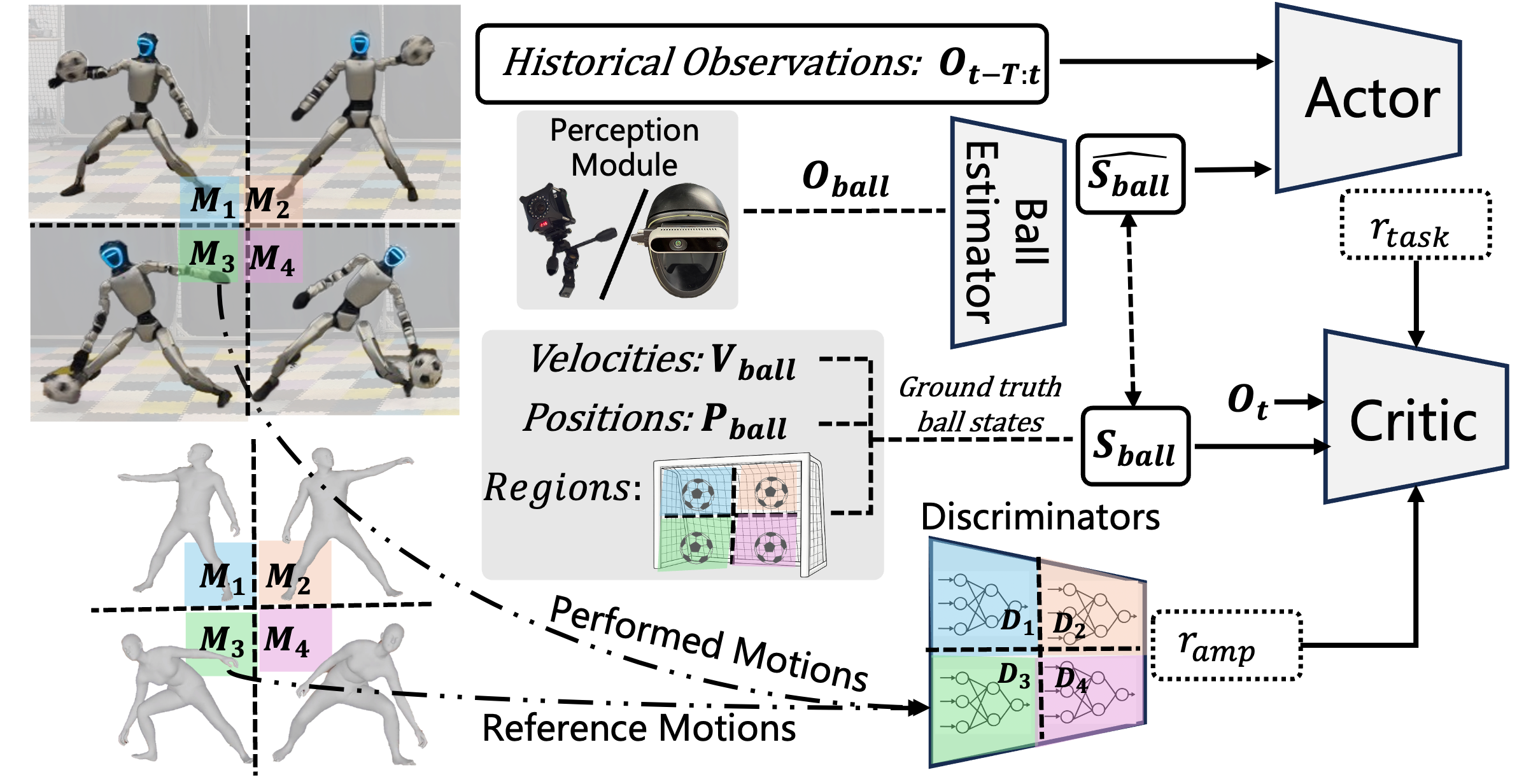} 
  \caption{Method framework: We train our policy using an end-to-end reinforcement learning (RL) framework, incorporating position-conditioned task and motion constraints. Our method is capable of operating with either an onboard camera or the motion-capture (MoCap) system.}
  \label{fig:method}
\end{figure}

We create the humanoid goalkeeper that can autonomously perform successfully keeping skills while adaptively conforming to human motion priors. Our approach (\cref{fig:method}) centers on an end-to-end reinforcement learning (RL) policy that generates highly dynamic and fully autonomous keeping behaviours. The policy operates directly on real-time ball position observations, which also serve as conditioning variables for integrated human motion priors. These priors are incorporated into the RL process via an adversarial training scheme, promoting robust and naturalistic motion generation without reliance on predefined trajectories or strict motion tracking.

Our method advances prior work by addressing two critical challenges: (1) We propose a unified framework that jointly enforces task success and motion realism through observation-conditioned adversarial training, eliminating the need for separate optimization stages; and (2) we demonstrate hardware-feasible, high-dynamic performance on real-world humanoid robots.

\subsection{Goalkeeper Training through RL}
Our learning system is designed to optimize two primary objectives: (1) task success and (2) motion resemblance to human demonstrations. This section details our reinforcement learning–based optimization process for enabling successful goalkeeping skills. 

\subsubsection{Training Environment}
We train policies using the standard PPO algorithm \cite{schulman2017proximal}, implemented in the IsaacGym simulator \cite{rudin2022learning}. The policy trains the robot to stop a flying ball before hitting into the goal line.

The observation space includes the observed ball`s position $\bm{O_{ball}}$, combined with proprioceptive observations $\bm{O_p}$. These are used as actor inputs $\bm{O_t}$, with a history of length $T$ to capture temporal information necessary for predicting the ball’s motion trajectory. 

Each episode begins with a ball launched toward the robot. The training environment is divided into $k$ regions, denoted as $\mathcal{R} = {0, 1, 2, \dots, k}$, where each region corresponds to a specific area on the goal line where the ball may arrive. The region is determined at the start of each episode, then we sample a random landing position $\mathbf{p}_\text{land}$ within that region and assign the ball's velocity accordingly to reach the target. \Cref{tab:obs_spaces} outlines the observation space used during training. All ball-related observations are computed in the robot’s local frame, facilitating direct deployment with onboard perception.
\begin{table}[t]
\caption{Observation spaces for the actor and critic network.}
\centering
\renewcommand{\arraystretch}{1.1}
\resizebox{0.5\textwidth}{!}{
\begin{tabular}{lcc}
\hline
\textbf{Observation} & \textbf{Actor} & \textbf{Critic} \\
\hline
Ball position (in local frame) ($\mathbf{O}_\text{ball} \in \mathbb{R}^3$) & \checkmark & \checkmark \\
Base angular velocity ($\boldsymbol{\omega}_{\text{base}} \in \mathbb{R}^3$) & \checkmark & \checkmark \\
Projected gravity vector ($\mathbf{g}_{\text{base}} \in \mathbb{R}^3$) & \checkmark & \checkmark \\
Joint positions ($\mathbf{q} \in \mathbb{R}^{29}$) & \checkmark & \checkmark \\
Joint velocities ($\dot{\mathbf{q}} \in \mathbb{R}^{29}$) & \checkmark & \checkmark \\
Previous action ($\mathbf{a}_{\text{last}} \in \mathbb{R}^{29}$) & \checkmark & \checkmark \\
Base linear velocity ($\mathbf{v}_{\text{base}} \in \mathbb{R}^3$) & -- & \checkmark \\
End target region ($\mathcal{R} = {0, 1, 2, \dots, k}$) & -- & \checkmark \\
End effector target ($\mathbf{p}_\text{target} \in \mathbb{R}^3$) & -- & \checkmark \\
Ball velocity (in local frame) ($\mathbf{v}_\text{ball} \in \mathbb{R}^3$) & -- & \checkmark \\
Left hand position ($\mathbf{p}_\text{hand}^\text{left}\in \mathbb{R}^3$) & -- & \checkmark \\
Right hand position ($\mathbf{p}_\text{hand}^\text{right}\in \mathbb{R}^3$) & -- & \checkmark \\
Reach Distance ($\|\mathbf{p}_\text{target} - \mathbf{p}_\text{hand}^{(\mathcal{R})}\|$) & -- & \checkmark \\
\hline
\end{tabular}}
\label{tab:obs_spaces}
\end{table}

\subsubsection{Position-Conditioned Task Rewards}

\begin{figure}[htbp]
  \centering
  \includegraphics[width=0.5 \textwidth]{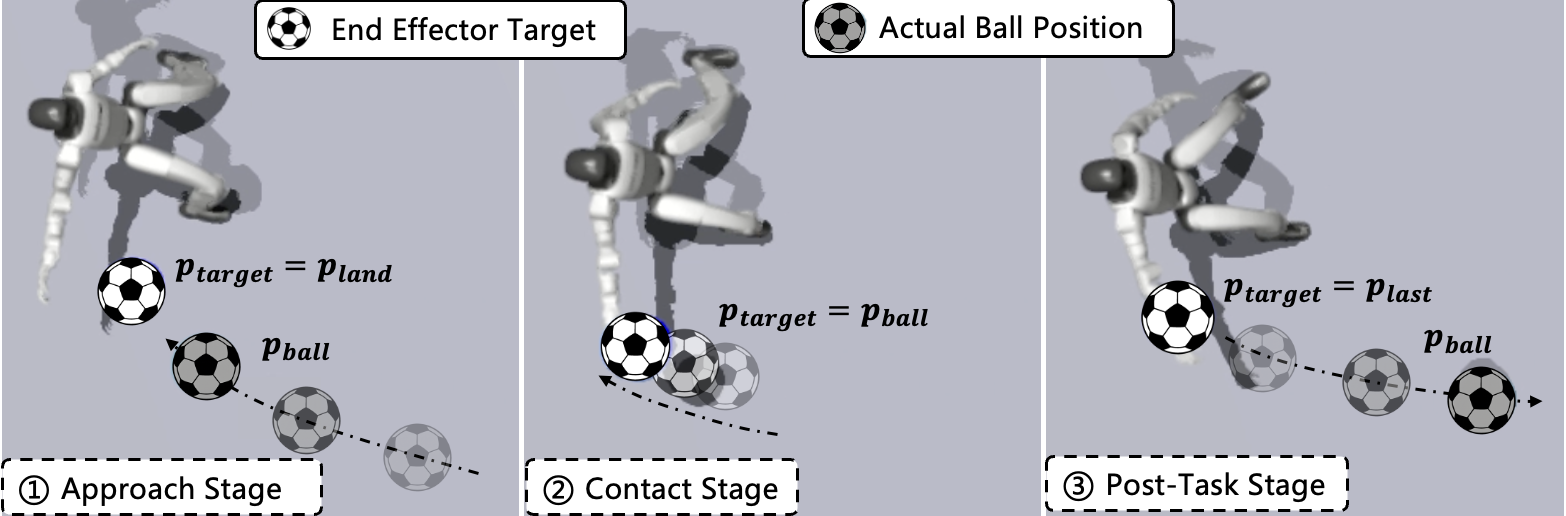} 
  \caption{This design ensures that when the ball is far from the robot, the interaction is based on the predicted landing point, whereas for closer balls, the target position is the actual ball location, allowing for more precise handling near the robot.}
  \label{fig:endtarget}
\end{figure}

We introduce \textbf{position-conditioned task rewards} $r_t$ that guide the robot to keep the ball away from the goal line by tracking a dynamic target. The end target $\mathbf{p}_\text{target}$ is designed to prompt the robot to predict the landing point of the approaching ball, touch and punch it away when it is close, and then hold a static position after the ball is stopped (\cref{fig:endtarget}). When the ball is flying towards the rbobot, we have:
\begin{equation}
\mathbf{p}_\text{target} = \mathbf{p}_\text{land} \cdot \mathbf{1}_{d_x > d_{th}} + \mathbf{p}_\text{ball} \cdot \mathbf{1}_{d_x \leq d_{th}},    
\end{equation}
where $\mathbf{1}_{\text{condition}}$ selects the corresponding target based on the distance between the ball and the robot, $d_x$ denotes the distance between the ball and the goal line, and $d_{th}$ stands for the approaching threhold. Then we define the following smooth, distance-based reward using a sigmoid function to encourage the end-effector (hand) to reach the target position:
\begin{equation}
\scriptsize
r_\text{t} = \{1 - \frac{1}{1 + \exp[-\sigma_\text{keep} (\|\mathbf{p}_\text{target} - \mathbf{p}_\text{hand}^{(\mathcal{R})}\| - d_\text{keep})]}\} \times (1 + \nu^{(\mathcal{R})}),
\end{equation}
where $\sigma_\text{keep}$ is the scaling factor and $d_\text{keep}$ is a catch distance threshold indicating when the end-effector is sufficiently close to the ball. The reward $r_t$ is conditioned on the landing region $\mathcal{R}$ through computing the distance between the selected hand and the end target $\|\mathbf{p}_\text{target} - \mathbf{p}_\text{hand}^{(\mathcal{R})}\|$. For example, if the ball is flying toward the left region, the robot should use its left hand to intercept it, and similarly for the right region.

We introduce a \textbf{position-conditioned dynamics modulation term} $\nu^{(\mathcal{R})}$ to encourage whole-body motions such as lateral shifts or jumps, depending on the region $\mathcal{R}$:
\begin{equation}
\nu^{(\mathcal{R})} = \nu_y \cdot \mathbf{1}_{\mathcal{R} \in \text{right}} + (-\nu_y) \cdot \mathbf{1}_{\mathcal{R} \in \text{left}} + \nu_z \cdot \mathbf{1}_{\mathcal{R} \in \text{up}},
\end{equation}
$\nu^{(\mathcal{R})}$ promotes consistent whole-body motions that support coordinated end-effector movements during interception.

In addition to the position-conditioned reward $r_t$, we incorporate additional reward terms to enforce motion constraints and promote hardware-feasible behavior. The full reward formulation is provided in the appendix.

\subsubsection{Post-Task Stablility}
\label{sec:post-task}
We set the episode length to 3 seconds, which is considerably longer than the ball's flight time (typically 0.4– seconds), in order to allow for post-task stabilization. To encourage the robot to maintain a normalized posture after the ball has passed, we introduce post-task reward terms that promote balance and recovery from the high-dynamic keeping motion. These rewards are generally defined as:
\begin{equation}
r_\text{post} = \exp\left(-\sigma_\text{stable} \cdot \| \mathbf{Err}_\text{stable} \|\right) \cdot \mathbf{1}_\text{ball stopped},
\end{equation}
where $\mathbf{Err}_\text{stable}$ denotes the deviation of key stability-related quantities—such as base height, angular velocity, and orientation—from their nominal values. The indicator function $\mathbf{1}_\text{ball stopped}$ activates the reward only after the ball has been kept or flying behind the robot. In addition, we reset a subset of environments not to the default joint configuration, but instead to the joint positions sampled from other ongoing environments. This forces the robot to begin a new trial in the middle of a previous keeping motion, rather than from a nominal reset state. In addition to promoting stability, we found this strategy effective for learning continuous keeping behaviors, as the robot can initiate new episodes without returning to a default pose.

\begin{figure*}[htbp]
  \centering
  \includegraphics[width=\textwidth]{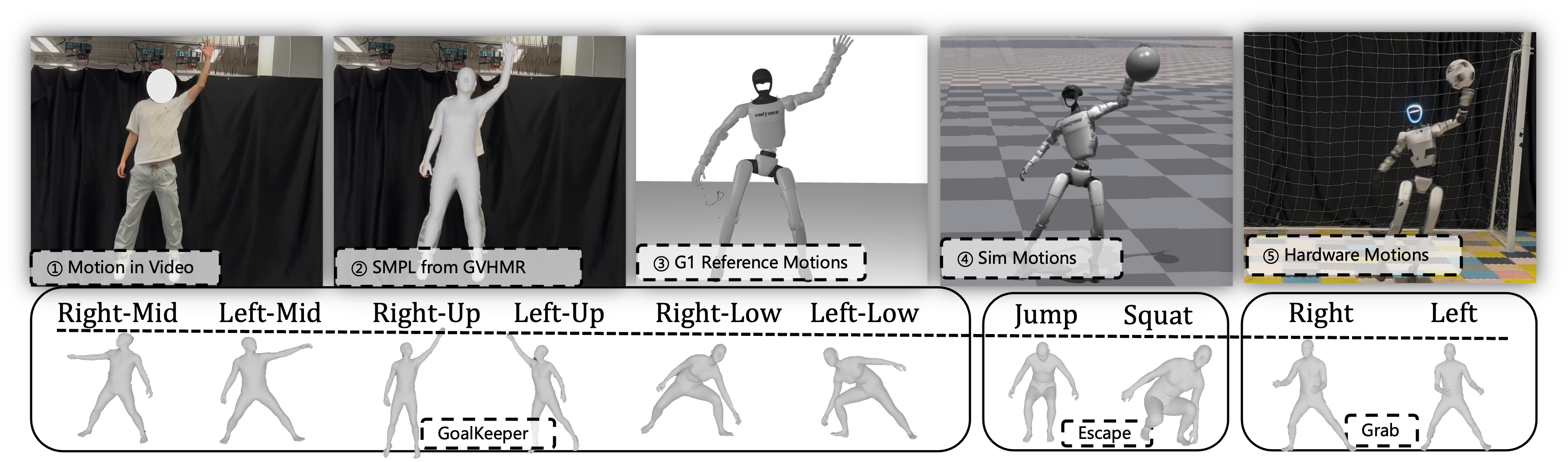} 
  \caption{The motion curation and formulation pipeline, demonstrating the process from human video to robot-executed motion.}
  \label{fig:gvhmr}
\end{figure*}

\subsection{Optimization with Motion Constraints}

We include a position-conditioned adversarial motion prior (AMP) reward to impose motion constraints during training, encouraging the robot to exhibit appropriate movement styles based on the ball's landing region (see \cref{fig:method}). Inspired by \cite{huang2023creating}, which addresses region-specific goalkeeping in quadrupeds, we incorporate imitation learning into reinforcement learning to promote region-consistent behaviors. These motion constraints are derived from a task-specific motion library constructed from human demonstrations.

\subsubsection{Reference Motion Curation}

We first leverage existing pipelines (GVHMR \cite{shen2024world})  that extract humanoid motions from self-recorded RGB videos (\cref{fig:gvhmr}). Next, we retarget the extracted motions to the Unitree G1 robot and store the resulting joint position sequences $\bm{q}_t$ in a region-specific motion reference buffer.

\subsubsection{Optimize with Motion Constraints}

The motion constraints are implemented using adversarial motion prior (AMP) rewards \cite{peng2021amp}, which encourage the robot to exhibit region-consistent motion styles during interception. To this end, we define position-conditioned AMP rewards by associating each region $\mathcal{R}$ with a dedicated reference motion slot and a corresponding discriminator $D^{(\mathcal{R})}$. Each discriminator is trained to distinguish between state transitions from the reference motions and those generated by the policy. The training objective for $D^{(\mathcal{R})}$ is:
\begin{align}
\scriptsize
\arg\min_{D^{(\mathcal{R})}} \quad 
& \mathbb{E}_{(q_t, q_{t+1}) \sim d_M^{(\mathcal{R})}} 
\left[(D^{(\mathcal{R})}(q_t, q_{t+1}) - 1)^2\right] \notag \\
& + \mathbb{E}_{(q_t, q_{t+1}) \sim d_\pi^{(\mathcal{R})}} 
\left[(D^{(\mathcal{R})}(q_t, q_{t+1}) + 1)^2\right],
\end{align}
where $d_M^{(\mathcal{R})}$ and $d_\pi^{(\mathcal{R})}$ denote the distributions of joint position transitions $(q_t, q_{t+1})$ from expert reference motions and from the policy, respectively, conditioned on region $\mathcal{R}$.

Given the high accuracy requirements of our task, we apply motion constraints more softly to avoid conflicts with task-specific objectives. Unlike the original AMP formulation—which directly scores executed motions and rewards those that closely match the reference distribution—we generate Gaussian samples around each executed motion and reward only the one with the highest discriminator score. This encourages smooth alignment with the reference motion distribution, allowing the policy to receive similar rewards for nearby motions while letting the task objectives determine the precise execution. The AMP reward is computed as:
\begin{equation}
\small
r_\text{amp} = \max_{j \in \{1, \dots, N\}} 
\left\{ 
\max \left[ 0, 1 - 0.25 \left(D(\tilde{q}^{(j)}_t, \tilde{q}^{(j)}_{t+1}) - 1\right)^2 
\right] 
\right\},
\end{equation}
where each $(\tilde{q}^{(j)}_t, \tilde{q}^{(j)}_{t+1})$ is sampled from a Gaussian centered at the executed transition $(q_t, q_{t+1})$.

\subsection{Aligning Sim2Real Perception}
Acquiring hardware-deployable ball perception and bridging the sim-to-real gap are critical challenges in dynamic object interaction tasks. In this work, the robot receives the ball position relative to its torso frame, represented by the observation vector $\bm{O}_{\text{ball}} \in \mathbb{R}^3$. We implement two hardware-based perception systems to acquire the ball position $\bm{O}_{\text{ball}}$: 1) a motion capture (MoCap) system using markers placed on the robot’s head and the ball, where $\bm{O}_{\text{ball}}$ is computed from the relative position between the two as reported by the MoCap system; 2) a depth camera (Intel RealSense D435i) mounted on the robot’s head with an infrared (IR) filter \cite{su2025toward}, which detects a high-reflectivity ball and directly estimates $\bm{O}_{\text{ball}}$ from the camera’s perception pipeline. While both systems provide reasonably accurate ball position estimates, performance gaps remain due to calibration errors and perception limitations—such as occlusions in the MoCap system and field-of-view constraints in the camera-based system. To mitigate these issues, we incorporate a ball position estimator into the policy and introduce training-time noise to improve robustness across both perception modalities.

\subsubsection{Ball Estimator}
We integrate a ball estimator into the training loop, which is effective in identifying regions (for task and motion conditions) and accurately predicting ball positions and moving trajectories (\cref{fig:method}), fully utilizing historical observations. The position estimator is trained with Mean Squared Error (MSE) loss to minimize the prediction error of the ball's position, while the region estimator is trained using Cross-Entropy Loss to classify the correct region where the ball is likely to land.

\subsubsection{Training Noises}
To simulate real-world perception uncertainties, we introduce noise into the ball observation $\bm{O}_{\text{ball}}$ during training. Specifically, we apply: 1) position noise—random perturbations of up to $5cm$, which is half the ball's $10cm$ radius—to mimic typical sensing errors; and 2) random dropouts—$\bm{O}_{\text{ball}}$ is occasionally set to zero after $0.4s$ of flight, simulating occlusions or out-of-FOV cases encountered in hardware perception when the ball approaches. Additionally, once the ball has stopped flying (e.g. after being intercepted or landing), we set $\bm{O}_{\text{ball}} = \mathbf{0}$ to reflect the absence of further tracking. The robot is then encouraged to hold its final pose, as described in \cref{sec:post-task}.
\section{Evaluations}

\begin{table*}[htbp]
  \centering
  \caption{Simulation Results.}
  \label{tab:Simulation Results}
  \resizebox{\textwidth}{!}{%
  \begin{tabular}{lcccccccc}
    \toprule
    \multirow{2}{*}{\textbf{Method}} &
      \multicolumn{2}{c}{\textbf{Task}} &
      \multicolumn{3}{c}{\textbf{Motion}} &
      \multicolumn{2}{c}{\textbf{Smoothness}}\\
    \cmidrule(lr){2-3}\cmidrule(lr){4-6} \cmidrule(lr){7-8}
     & $E_{\text{succ}}\!\uparrow$ & $E_{\text{ee task}}\!\downarrow$ & $E_{\text{match}(\mathcal{R})}\!\uparrow$ &  $E_{\text{pos}(min)}\!\downarrow$  &  $E_{\text{pos}(\mathcal{R})}\!\downarrow$ & $E_{\text{dof acc}}\!\downarrow$ & $E_{\text{smth}}\!\downarrow$ \\
    \midrule
    \rowcolor{gray!18}\multicolumn{8}{l}{\textbf{Ablation Motion / Task Constraints}} \\
    w.o. Task Constraints   & $9.44\!\pm\! \mathsmaller{4.82}$ & $0.55\!\pm\! \mathsmaller{0.018}$ & $64.33\!\pm\! \mathsmaller{5.37}$ & $\bm{0.76}\!\pm\! \mathsmaller{0.012}$ &  $\bm{0.98}\!\pm\! \mathsmaller{0.060}$ & $\bm{14.26}\!\pm\! \mathsmaller{0.275}$ & $\bm{0.030}\!\pm\! \mathsmaller{0.001}$ \\
    w.o. Motion Constraints    & $31.69\!\pm\! \mathsmaller{1.18}$ & $0.98\!\pm\! \mathsmaller{0.172}$ & $8.83\!\pm\! \mathsmaller{4.50}$ & $1.73\!\pm\! \mathsmaller{0.018}$ &  $2.03\!\pm\! \mathsmaller{0.025}$ & $62.23\!\pm\! \mathsmaller{4.368}$ & $0.151\!\pm\! \mathsmaller{0.015}$ \\
    Humanoid Goalkeeper (ours) & $\bm{80.92}\!\pm\! \mathsmaller{1.72}$ & $\bm{0.12}\!\pm\! \mathsmaller{0.004}$ & $\bm{67.84}\!\pm\! \mathsmaller{1.49}$ & $1.24\!\pm\! \mathsmaller{0.017}$ & $1.31\!\pm\! \mathsmaller{0.021}$ & $22.80\!\pm\! \mathsmaller{0.404}$ & $0.042\!\pm\! \mathsmaller{0.001}$ \\
    \midrule
    \rowcolor{gray!18}\multicolumn{8}{l}{\textbf{Ablation Position Division}} \\
    w.o.\ AMP Division   & $76.11\!\pm\! \mathsmaller{2.19}$ & $0.14\!\pm\! \mathsmaller{0.010}$ & $25.73\!\pm\! \mathsmaller{5.03}$ & $1.39\!\pm\! \mathsmaller{0.007}$ & $1.51\!\pm\! \mathsmaller{0.015}$ & $23.18\!\pm\! \mathsmaller{0.326}$ & $0.043\!\pm\! \mathsmaller{0.000}$\\
    w.o.\ Task Division  & $62.08\!\pm\! \mathsmaller{4.30}$ & $0.14\!\pm\! \mathsmaller{0.005}$ & $39.18\!\pm\! \mathsmaller{3.60}$ & $1.62\!\pm\! \mathsmaller{0.007}$ & $1.84\!\pm\! \mathsmaller{0.030}$ & $28.69\!\pm\! \mathsmaller{0.329}$ & $0.054\!\pm\! \mathsmaller{0.000}$\\
    w.o.\ Division       & $64.83\!\pm\! \mathsmaller{5.07}$ & $0.14\!\pm\! \mathsmaller{0.004}$ & $19.09\!\pm\! \mathsmaller{5.39}$ & $1.62\!\pm\! \mathsmaller{0.023}$ & $1.87\!\pm\! \mathsmaller{0.057}$ & $28.56\!\pm\! \mathsmaller{0.450}$ & $0.052\!\pm\! \mathsmaller{0.001}$\\
    Humanoid Goalkeeper (ours)& $\bm{80.92}\!\pm\! \mathsmaller{1.72}$ & $\bm{0.12}\!\pm\! \mathsmaller{0.004}$ & $\bm{67.84}\!\pm\! \mathsmaller{1.49}$ & $\bm{1.24}\!\pm\! \mathsmaller{0.017}$ & $\bm{1.31}\!\pm\! \mathsmaller{0.021}$ & $\bm{22.80}\!\pm\! \mathsmaller{0.404}$ & $\bm{0.042}\!\pm\! \mathsmaller{0.001}$\\
    \midrule
    \rowcolor{gray!18}\multicolumn{8}{l}{\textbf{Evaluation Settings}} \\
    Speed-Easy $(0.7s \sim 1.0s)$ & $80.95\!\pm\! \mathsmaller{4.21}$ & $0.12\!\pm\! \mathsmaller{0.012}$ & $50.88\!\pm\! \mathsmaller{3.28}$ & $1.26\!\pm\! \mathsmaller{0.008}$ & $1.37\!\pm\! \mathsmaller{0.021}$ & $23.20\!\pm\! \mathsmaller{0.432}$ & $0.042\!\pm\! \mathsmaller{0.001}$ \\
    Range-Easy $(\pm 1.0m)$ & $\bm{84.64}\!\pm\! \mathsmaller{2.75}$ & $0.13\!\pm\! \mathsmaller{0.011}$ & $63.16\!\pm\! \mathsmaller{4.47}$ & $1.26\!\pm\! \mathsmaller{0.007}$ & $1.32\!\pm\! \mathsmaller{0.012}$ & $22.23\!\pm\! \mathsmaller{0.244}$ & $0.041\!\pm\! \mathsmaller{0.000}$ \\
    Speed-Hard $(0.4s \sim 0.7s)$ & $64.25\!\pm\! \mathsmaller{7.23}$ & $0.13\!\pm\! \mathsmaller{0.009}$ & $\bm{70.47}\!\pm\! \mathsmaller{3.53}$ & $1.26\!\pm\! \mathsmaller{0.013}$ & $\bm{1.29}\!\pm\! \mathsmaller{0.015}$ & $\bm{21.85}\!\pm\! \mathsmaller{0.324}$ & $\bm{0.041}\!\pm\! \mathsmaller{0.000}$ \\
    Range-Hard $(\pm 2.0m)$ & $63.91\!\pm\! \mathsmaller{2.29}$ & $0.15\!\pm\! \mathsmaller{0.007}$ & $63.45\!\pm\! \mathsmaller{7.92}$ & $1.26\!\pm\! \mathsmaller{0.009}$ & $1.33\!\pm\! \mathsmaller{0.012}$ & $24.42\!\pm\! \mathsmaller{0.449}$ & $0.044\!\pm\! \mathsmaller{0.001}$ \\
    Default ($0.5s \sim 1.0s, \quad \pm 1.5m)$  &  $80.92\!\pm\! \mathsmaller{1.72}$ & $\bm{0.12}\!\pm\! \mathsmaller{0.004}$ & $67.84\!\pm\! \mathsmaller{1.49}$ & $\bm{1.24}\!\pm\! \mathsmaller{0.017}$ & $1.31\!\pm\! \mathsmaller{0.021}$ & $22.80\!\pm\! \mathsmaller{0.404}$ & $0.042\!\pm\! \mathsmaller{0.001}$\\
    \bottomrule
  \end{tabular}}
\end{table*}

\begin{figure*}[htbp]
  \centering
  \includegraphics[width=\textwidth]{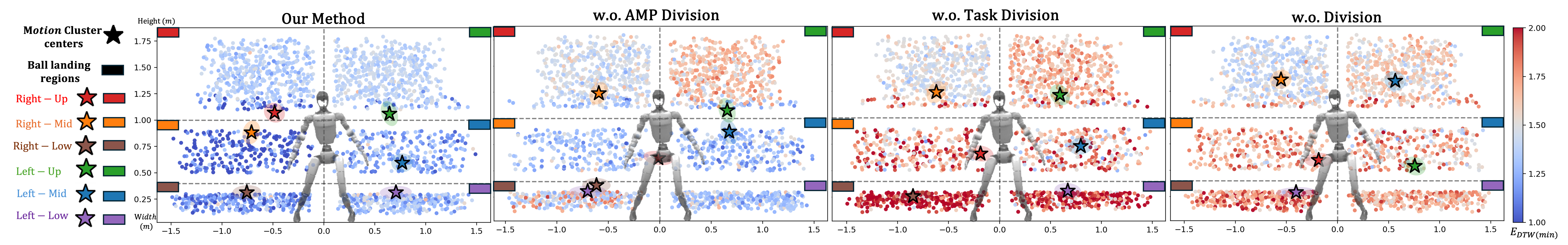} 
  \caption{Each scatter point represents a trial, with the position corresponding to the ball target of the trial. The color gradient of the points indicates $E_{\text{pos}}(\text{min})$ for each trial. The pentagram represents the cluster center of trials that has the same matching region (regions with minimal $E_{\text{pos}})$. A well-learned position-conditioned policy should exhibit a clear correspondence between the cluster centers and their respective ball target regions}
  \label{fig:simvisual}
\end{figure*}

We conduct a series of experiments to comprehensively evaluate the proposed method, focusing on the following objectives:  
1) to validate the effectiveness of our method in mastering humanoid goalkeeping skills (in terms of success rate and coverage),  
2) to assess the resemblance between robot-executed motions and expert demonstrations, and  
3) to evaluate the benefit of the proposed position-conditioned learning scheme compared to non-division baselines. 
To this end, we design evaluation metrics that reflect task performance, motion fidelity, and action smoothness:
\newcolumntype{L}[1]{>{\raggedright\arraybackslash}p{#1}}
\begin{center}
\renewcommand{\arraystretch}{1.3}
\rowcolors{1}{gray!8}{white}
\resizebox{0.48\textwidth}{!}{
\begin{tabular}{@{}cL{6.8cm}@{}}
\toprule
$E_{\text{succ}}\!\uparrow$ & Success rate of blocking the ball. \\
$E_{\text{ee task}}\!\downarrow$ & Closest distance between the end effector (hand) and ball during a trial. \\
$E_{\text{pos}(\mathcal{R})}\!\downarrow$ & pose error to expert motion (processed through DTW \cite{sakoe2003dynamic}) from the ball target region $\mathcal{R}$. \\
$E_{\text{pos}(min)}\!\downarrow$ & Lowest pose error between executed motion and expert motion among all the regions. \\
$E_{\text{match}(\mathcal{R})}\!\uparrow$ & Whether the closest-matching motion aligns with the ball target region $\mathcal{R}$. \\
$E_{\text{dof acc}}\!\downarrow$ & Joint acceleration: $\frac{1}{\Delta t} \left| \dot{q}_{t} - \dot{q}_{t-1} \right|$. \\
$E_{\text{smth}}\!\downarrow$ & Action smoothness: $\left| a_t - 2a_{t-1} + a_{t-2} \right|$. \\
\bottomrule
\end{tabular}}
\end{center}

During training and evaluation in simulation, the goal area is partitioned into six predefined ball landing regions, with corresponding expert motions visualized in \cref{fig:gvhmr} and the region boundaries illustrated in \cref{fig:simvisual}. We conduct 500 simulation trials and 5 hardware trials for each region. The ball flight duration ranges from $0.5\,\mathrm{s}$ to $1.0\,\mathrm{s}$, with flying distances varying from $3.0\,\mathrm{m}$ to $5.0\,\mathrm{m}$. 
The goal used in hardware evaluations corresponds to the standard dimensions of a 5-a-side soccer match, with a width of $3.0\,\text{m}$ (between the goalposts) and a height of $2.0\,\text{m}$.

\subsection{Simulation Results}
\cref{tab:Simulation Results} summarizes the primary numerical results from our simulation experiments. The proposed method achieves the highest task success rate while also demonstrating superior motion resemblance compared to all baselines. These outcomes support our core hypothesis: aligning task constraints with appropriate motion constraints yields better overall performance. Notably, the \textit{w.o.\ AMP Division} variant—which lacks explicit motion segmentation but retains region-based task constraints—achieves moderately competitive success and motion metrics. This result suggests that correct task conditioning can guide action selection to some extent. However, its performance still lags behind the full method, highlighting the importance of explicit motion correspondence in achieving high-quality, task-aligned behaviors.

Given the default keeping ranges of $(\pm 1.5\,\text{m})$ in width and $(1.8\,\text{m})$ in height, the proposed method achieves a notably high success rate—despite operating over one of the widest goalkeeping areas reported in the robotic goalkeeping literature. To further assess robustness, we evaluate the method under varying ball speeds and expanded guarding regions. As expected, the results indicate a gradual decline in performance as task difficulty increases, validating the method’s scalability while revealing its current limitations under more challenging conditions.

The smoothness metric serves as a useful indicator for assessing how well the alignment between motion and task constraints translates into natural robot behavior. Our results support this connection: our policy that guided by accurate expert motions achieve better smoothness. Interestingly, the baseline without explicit task constraints—relying solely on AMP and regularization rewards—achieves the best smoothness performance. This suggests that, in the absence of region-based goals, the robot may prioritize generic motion regularity over task effectiveness, resulting in smoother but less purposeful behaviors.

\cref{fig:simvisual} provides a region-based visualization of motion behavior to validate our approach. In each subplot, the pentagram represents the cluster center of robot-performed motions that best match the specific expert motion. The results show that our method learns region-conditioned motion effectively, with distinct and consistent clusters emerging for each region. In contrast, baseline methods tend to mix motions across regions, resulting in less coherent behaviors and higher joint position errors. This highlights the effectiveness of our region-conditioned policy learning in preserving motion structure aligned with task semantics.

\begin{table*}[htbp]
  \centering
  \caption{Hardware Results.}
  \label{tab:Hardware Results}
  \resizebox{\textwidth}{!}{%
  \begin{tabular}{lccc ccc ccc}
    \toprule
    \multirow{2}{*}{\textbf{Method}} &
      \multicolumn{3}{c}{\textbf{Right-Low}} &
      \multicolumn{3}{c}{\textbf{Right-Mid}} &
      \multicolumn{3}{c}{\textbf{Right-Up}} \\
    \cmidrule(lr){2-4}\cmidrule(lr){5-7} \cmidrule(lr){8-10}
     & $E_{\text{succ}}\!\uparrow$ & $E_{\text{pos}(min)}\!$ & $E_{\text{dof acc}}\!\downarrow$ 
     & $E_{\text{succ}}$ & $E_{\text{pos}(min)}\!$ & $E_{\text{dof acc}}$
     & $E_{\text{succ}}$ & $E_{\text{pos}(min)}\!$ & $E_{\text{dof acc}}$\\
    \midrule
    Humanoid Goalkeeper (camera)
    & $1/5$      & $1.67$      & $\bm{21.79}$  & $3/5$ & $1.46$ & $21.30$  & $\bm{3/5}$ & $1.56$ & $32.99$ \\  
    w.o.\ AMP Division (mocap)      
    & $0/5$      & $\bm{1.43}$ & $31.02$  & $3/5$ & $1.36$ & $\bm{20.62}$ & $0/5$ & $1.47$ & $31.33$ \\
    w.o.\ Division (mocap)       
    & $1/5$      & $1.66$      & $36.13$  & $0/5$ & $1.46$ & $31.09$ & $2/5$ & $1.57$ & $34.47$ \\
    Humanoid Goalkeeper (mocap)
    & $\bm{5/5}$ & $1.49$      & $33.34$  & $\bm{4/5}$ & $\bm{1.28}$ & $24.56$ & $1/5$ & $\bm{1.43}$ & $\bm{29.43}$ \\
    \midrule
    \multirow{2}{*}{\textbf{Method}} &
      \multicolumn{3}{c}{\textbf{Left-Low}} &
      \multicolumn{3}{c}{\textbf{Left-Mid}} &
      \multicolumn{3}{c}{\textbf{Left-Up}}\\
    \cmidrule(lr){2-4}\cmidrule(lr){5-7} \cmidrule(lr){8-10}
     & $E_{\text{succ}}\!\uparrow$ & $E_{\text{pos}(min)}\!$ & $E_{\text{dof acc}}\!\downarrow$ 
     & $E_{\text{succ}}$ & $E_{\text{pos}(min)}\!$ & $E_{\text{dof acc}}$
     & $E_{\text{succ}}$ & $E_{\text{pos}(min)}\!$ & $E_{\text{dof acc}}$\\
    \midrule
    Humanoid Goalkeeper (camera)
                         & $1/5$ & $1.70$ & $\bm{24.04}$  & $3/5$ & $1.53$ & $21.71$  & $3/5$ & $1.58$ & $28.12$ \\
    w.o.\ AMP Division (mocap)   & $5/5$ & $\bm{1.48}$ & $35.25$  & $2/5$ & $1.54$ & $\bm{18.60}$ & $3/5$ & $1.47$ & $27.68$ \\
    w.o.\ Division (mocap)       & $3/5$ & $1.48$ & $29.88$  & $3/5$ & $\bm{1.43}$ & $23.77$ & $3/5$ & $1.68$ & $38.87$\\
    Humanoid Goalkeeper (mocap)
                         & $5/5$ & $1.68$ & $\bm{28.76}$  & $\bm{3/5}$ & $1.44$ & $22.08$  & $3/5$ & $\bm{1.40}$ & $\bm{27.16}$ \\

    \bottomrule
  \end{tabular}}
\end{table*}

\begin{figure*}[htbp]
  \centering
  \includegraphics[width= \textwidth]{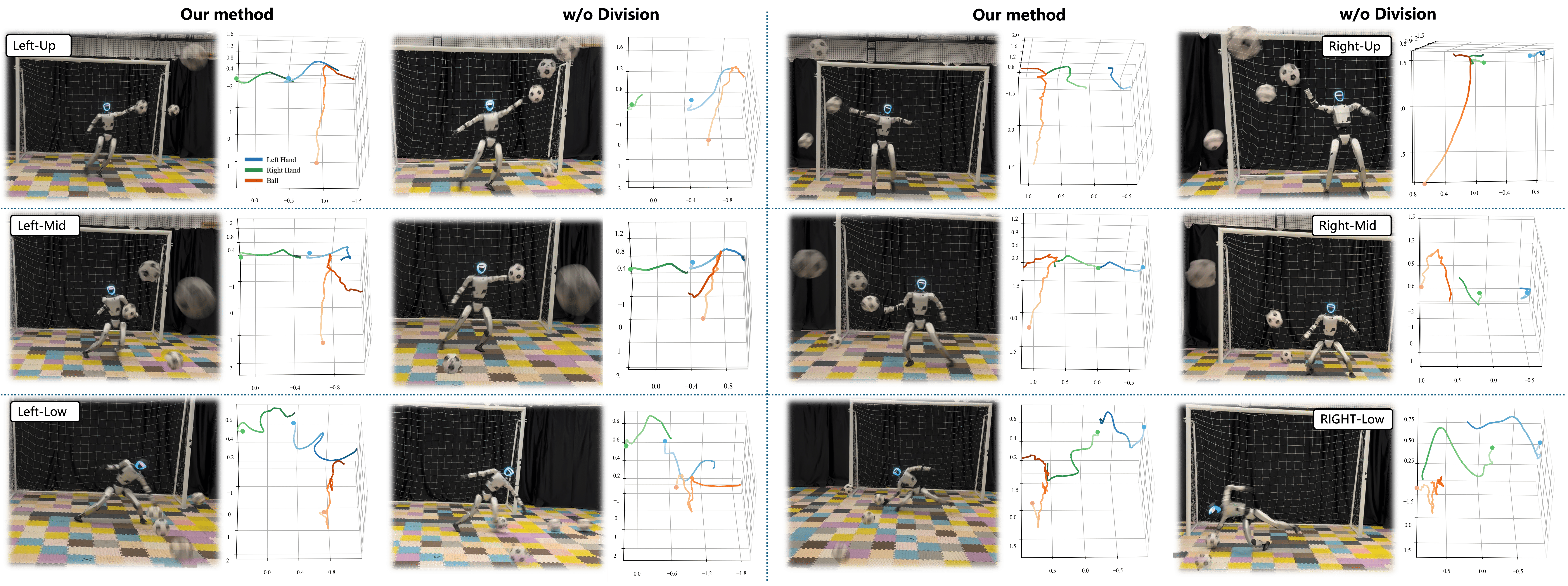} 
  \caption{We visualize the 3D trajectory of the ball, as well as the left and right hand trajectories for one trial in each region, and compare the performance of the proposed method with that of the non-division baselines.}
  \label{fig:trajvis}
\end{figure*}

\subsection{Region-Specific Hardware Evaluation}
\cref{tab:Hardware Results} presents region-specific hardware evaluation results on the goalkeeping task. The proposed method achieves a total success rate of $21/30$, with most failure cases concentrated in the upper regions. These upper regions demand stricter alignment between the end-effector and the ball trajectory, as well as wide-range, whole-body motions—posing greater challenges compared to the lower and side regions.

Despite these challenges, our method exhibits the smallest sim-to-real performance drop among all compared methods. In contrast, non-division baselines tend to overfit to one side of the goal and underperform in other regions. Specifically, their performance on the left side consistently exceeds that on the right, indicating a lack of generalization across the full goal area. We further illustrate this phenomenon in \cref{fig:trajvis}, which visualizes the ball trajectory and end-effector motion (recorded via MoCap). The visualization shows that, for balls targeting the right-side regions, the robot controlled by a non-division baseline fails to initiate an early lateral step, resulting in missed saves due to the short response window. Conversely, our method accurately guides the end-effector to the appropriate region from the onset, resulting in successful interception. This highlights the effectiveness of region-aware motion learning in achieving balanced, robust performance across the entire goal.

\begin{figure*}[htbp]
  \centering
  \includegraphics[width= \textwidth]{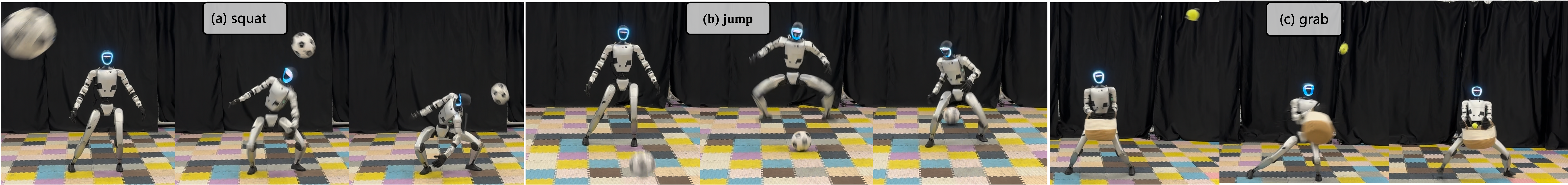} 
  \caption{We demonstrate the generalization ability of the extended tasks, including ball grabbing and ball escaping.}
  \label{fig:escape}
\end{figure*}

\begin{figure}[htbp]
  \centering
  \includegraphics[width= 0.5\textwidth]{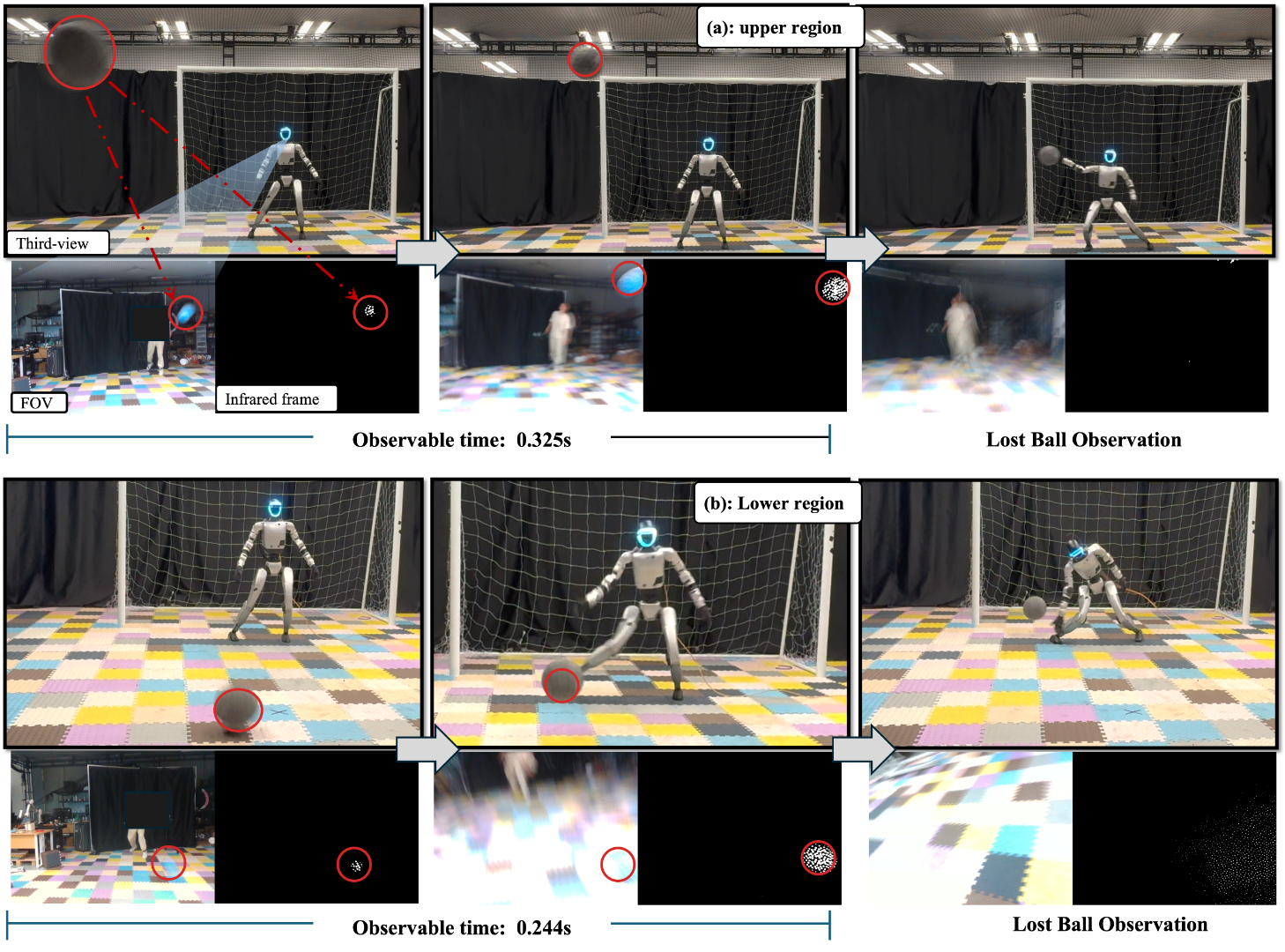} 
  \caption{the first-view illustration of on-board camera perception.}
  \label{fig:camera}
\end{figure}
\subsection{Performance with Camera perception}
The performance under camera-based perception drops to $14/30$ successful trials, compared to the MoCap setup. The primary limitation stems from the narrow camera field of view (FoV), which reduces the observable time window of the ball. As shown in \cref{fig:camera}, the high-reflectivity ball is easily detectable through the camera's infrared channel at a distance. However, as the ball approaches the robot, it quickly disappears from view, leaving less than $0.3$ seconds of observable time for lower regions. This limited visibility contributes to the decline in success rate due to the lack of consistent ball tracking.

Despite the FoV constraints, the camera-based system still achieves high success rates in the upper regions, where the ball remains visible for longer periods. This result demonstrates the robustness and adaptability of the proposed method when relying solely on onboard perception. Furthermore, the perceptual constraint naturally suppresses large, abrupt motions (such as lateral jumps), resulting in smoother behaviors, as reflected by improved $E_{\text{dof acc}}$ scores compared to the MoCap-based evaluations.

\subsection{Perform Continuous Keeping}

\begin{figure}[htbp]
  \centering
  \includegraphics[width= 0.5\textwidth]{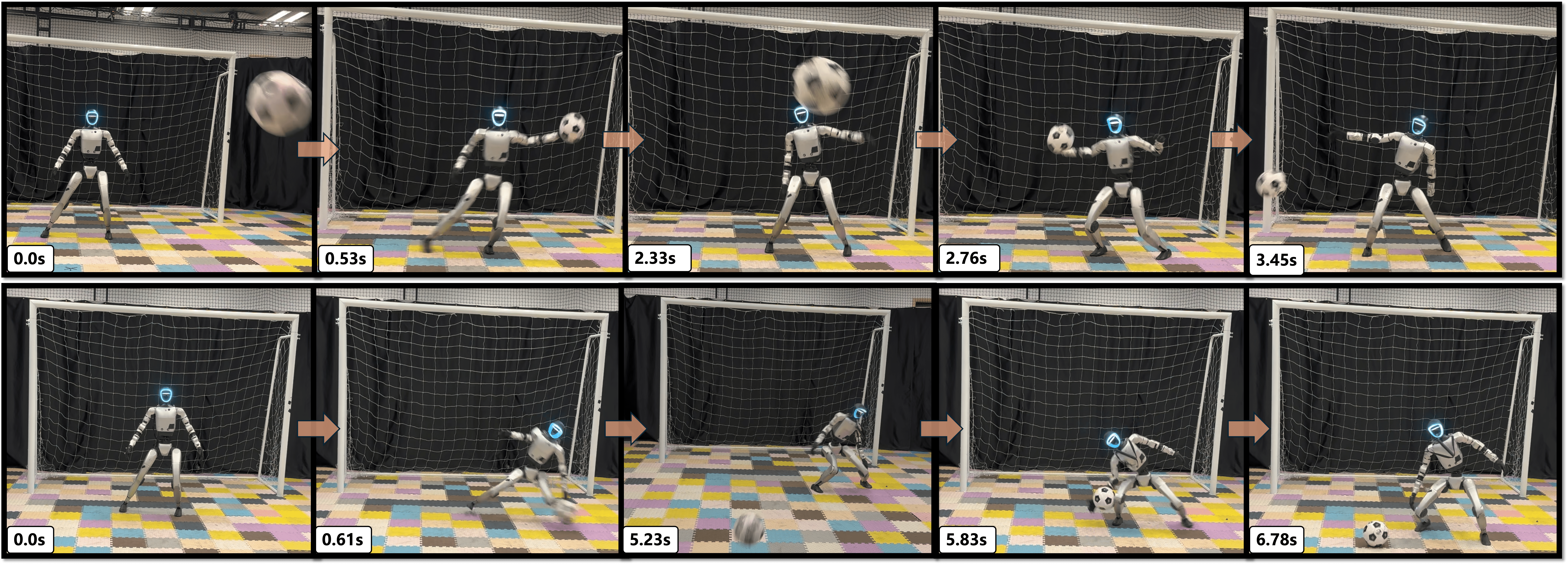} 
  \caption{Snapshots of the robot performing continuous goalkeeping without resetting to the default posture.}
  \label{fig:continue}
\end{figure}

\cref{fig:continue} demonstrates that the proposed pipeline enables the robot to continuously perform goalkeeping skills without requiring a manual reset to default states between trials. This is evidenced by two consecutive successful interceptions from both sides of the goal. As shown by the timestamps on the snapshots, the robot transitions directly from the terminal posture of the first trial to the initiation of the second, without entering a recovery phase. This capability highlights the potential of our method for real-world scenarios that demand uninterrupted, match-like behavior.

\subsection{Generalization Tasks}

Centered on the goalkeeper task, we introduce an effective framework that enables humanoid robots to perform whole-body interactions with dynamic objects by leveraging learned motion priors. To demonstrate the generalization capability of our framework beyond goalkeeping, we extend it to two additional tasks that also demand highly dynamic responses: (1) escaping from an incoming flying ball, and (2) grabbing a moving ball using a soft bag.

These motions are developed using the same position-conditioned approach employed in goalkeeping. Specifically, we partition the input space based on the ball's spatial characteristics: high vs. low positions correspond to jump and squat escapes, respectively; left vs. right positions correspond to left- and right-stepped ball grabs. We show this region-based division ensures appropriate motion selection under varying conditions. \cref{Escape} presents the quantitative results of our method on these tasks.

\begin{table}[!htbp]
\centering
\caption{Escape Task Evaluation.}
\setlength{\tabcolsep}{8pt}
\resizebox{0.9\linewidth}{!}{%
\begin{tabular}{lcccc}
\toprule[1.0pt]
Task & $E_{\text{succ}}\!\uparrow$ & $E_{\text{pos}(min)}\!$ & $E_{\text{dof acc}}\!\downarrow$ \\
\midrule[0.8pt]
Jump Escape   & $3/5$ & $1.49$ & $46.79$\\
Squat Escape  & $5/5$ & $1.12$ & $16.19$ \\ 
Grab Tennis   & $1/5$ & $0.99$ & $18.25$ \\
\bottomrule[1.0pt]
\end{tabular}}
\label{Escape}
\end{table}

\begin{figure}[htbp]
  \centering
  \includegraphics[width= 0.45\textwidth]{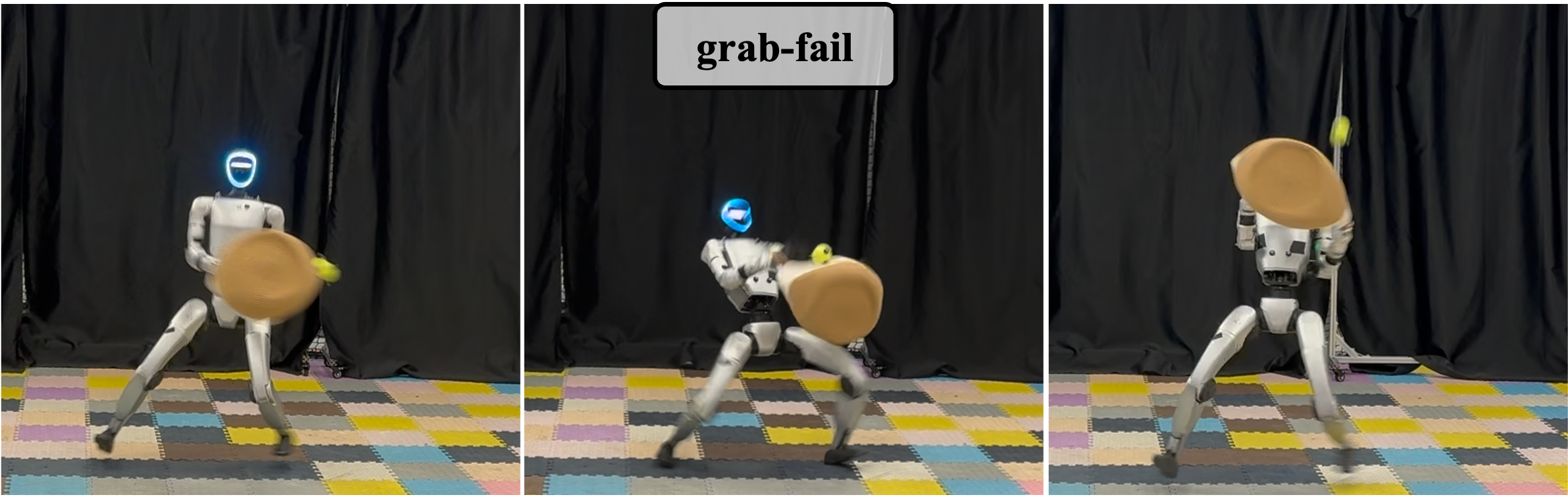} 
  \caption{The uncontrollable nature of the bag contributes to the low grabbing success rate.}
  \label{fig:grab fail}
\end{figure}

In the escape task, the robot achieves a high success rate and closely replicates expert motion \cref{fig:escape}, validating the effectiveness of our generalzaition ability. The robot is capable of reacting to an incoming ball and executing a successful escape maneuver within 0.5 seconds. Meanwhile, jumping from a stationary stance constitutes a highly dynamic maneuver, leading to substantial joint accelerations. The grabbing task exhibits a lower success rate, due to the absence of bag during training. In simulation, a successful grab is defined as the ball falling between the robot’s two hands, implicitly assuming a consistently open and rigid bag structure. However, in real-world trials, the deformability and varying orientation of the soft bag’s mouth frequently violate this assumption, as illustrated in \cref{fig:grab fail}. Despite these difficulties, we observe that the robot reliably moves toward the ball’s predicted landing location.

\section{discussion}

We have developed a humanoid robotic goalkeeper capable of executing agile, human-like motions to intercept flying balls, as well as performing tasks such as escaping a ball using jump and squat movements. Our results show that explicitly aligning task and motion constraints based on task observations significantly improves both motion quality and task completion. Extensive experimental evaluations, along with alternative perception modules, provide strong evidence supporting the effectiveness of our method.

However, as the first humanoid-level autonomous goalkeeper, there is still room for improvement before achieving professional-level goalkeeping performance. Specifically: 

(1) the lack of global observations and a high-level planning module limits our framework’s ability to perform consecutive goalkeeping actions at the match level.

(2) the perception module requires further refinement to react appropriately to complex situations in real-world scenarios. Currently, the robot responds to any ball detected within its field of view, even when the ball is not targeting the goal, which is not valid in real matches.

Future work could focus on adding a high-level planner to enable continuous goalkeeping actions, such as repositioning after each trial. Additionally, improving the perception system by integrating active sensing capabilities to overcome camera field-of-view (FOV) limitations and eliminating the reliance on fixed motion-capture systems will be crucial for enhancing robustness and flexibility in real-world settings. Furthermore, to advance the robot's current motions, we believe training it to grab the ball with its hand, rather than simply punching it away, would present an appealing challenge and a valuable demonstration for humanoid-object interaction skills.

\section{Acknowledgment}
This work was conducted during the author’s internship at the Embodied AI Center of Shanghai AI Laboratory. We would like to thank the Intelligent Photonics and Electronics Center at Shanghai AI Laboratory for supporting the motion-capture system and providing access to the experimental field. We are especially grateful to Yuman Gao for sharing the codebase for flying ball interception using an onboard camera, adapted from his previous work\cite{su2025toward}. This paper is partially supported by the National Key R\&D Program of China No.2022ZD0161000 and the General Research Fund of Hong Kong No.17208825.



\printbibliography

\end{document}